\documentclass[hidelinks, a4paper,twoside]{article}

\usepackage{epsfig}
\usepackage{subcaption}
\usepackage{calc}
\usepackage{amssymb}
\usepackage{amstext}
\usepackage{amsmath}
\usepackage{amsthm}
\usepackage{multicol}
\usepackage{pslatex}
\usepackage{apalike}
\usepackage{enumitem}
\usepackage{multirow}
\usepackage{spreadtab}
\usepackage{ragged2e}
\usepackage{breakcites}
\usepackage{etoolbox}
\usepackage{verbatim}
\usepackage{hyperref}
\makeatletter
\patchcmd{\@citex}{,}{;}{}{}
\makeatother
\usepackage{SCITEPRESS}     % Please add other packages that you may need BEFORE the SCITEPRESS.sty package.

\begin{document}
	
	\title{Hybrid Tiled Convolutional Neural Networks (HTCNN) \\  Text Sentiment Classification}
	
		%\author{Anonymous Authors 
		%\affiliation{\sup{1}Anonymous University}
		%\email{email@email.com}
	%}
	
%\begin{comment}
	\author{Maria Mihaela Tru\c{s}c\v{a}\sup{1}\orcidAuthor{0000-0000-0000-0000} and \authorname{Gerasimos Spanakis\sup{2}\orcidAuthor{0000-0002-0799-024}  }
		\affiliation{\sup{1}Department of Informatics and Economic Cybernetics, Bucharest University of Economic Studies, Bucharest, Romania}
		\affiliation{\sup{2}Department of Data Science and Knowledge Engineering, Maastricht University, Maastricht, The Netherlands}
		\email{maria.trusca@csie.ase.ro, jerry.spanakis@maastrichtuniversity.nl}
	}
%	\end{comment}

	\keywords{Hybrid Tiled Convolutional Neural Network, Sentiment Analysis.}
	
	\abstract{The tiled convolutional neural network (TCNN) has been applied only to computer vision for learning invariances. We adjust its architecture to NLP to improve the extraction of the most salient features for sentiment analysis. Knowing that the major drawback of the TCNN in the NLP field is its inflexible filter structure, we propose a novel architecture called hybrid tiled convolutional neural network (HTCNN) that applies a filter only on the words that appear in similar contexts and on their neighbouring words (a necessary step for preventing the loss of some $n$-grams). The experiments on the IMDB movie reviews dataset demonstrate the effectiveness of the HTCNN that has a higher level of performance of more than 3\% and 1\% respectively than both the convolutional neural network (CNN) and the TCNN. These results are confirmed by the SemEval-2017 dataset where the recall of the HTCNN model exceeds by more than six percentage points the recall of its simple variant, CNN.  
	}
	
	\onecolumn \maketitle \normalsize \setcounter{footnote}{0} \vfill
	
	\section{\uppercase{Introduction}}
	\label{sec:introduction}
	
	\noindent Sentiment analysis or opinion mining is a sort of text classification that assigns a sentiment orientation to documents based on the detected contextual polarity \cite{liu2012sentiment}. In the past, research work has focused only on the overall sentiment of a document, trying to determine if the entire text is positive, neutral or negative \cite{liu2012survey}. However, besides predicting the general sentiment, a better understanding of the reviews could be undertaken using the more in-depth aspect based sentiment analysis (ABSA) \cite{thet2010aspect, liu2012survey}. Specifically, ABSA’s goal is to predict the sentiment polarity of the corpus’ target entities and aspects (e.g. the possible aspects of the entity “movie” could be “the plot”, “the actors’ acting” or “the special effects”). While it is easy to establish the text entities based on the a priori information about text corpus, the aspect terms are difficult to infer and usually the training process requires to use a predefined set of term categories \cite{wang2015deep, zhang2012weakness, ma2018targeted}.
	
	In this paper, we propose a framework that applies the idea of ABSA only conceptually and tries to enhance the performance of one of the most popular baseline models for sentiment classification. More specifically, we adjust a convolutional neural network (CNN) to the ABSA requirements forcing it to capture more diverse features (both sentiment information and aspects). The CNN models were proposed for object recognition \cite{lecun1999object} and initially were used within computer vision. Soon they were adapted or integrated with other deep learning architectures to be compatible with a broad array of tasks, including NLP: classification \cite{lai2015recurrent}, recognition of entailment and contraction between sentences \cite{mou2015natural} or question answering \cite{feng2015applying}. 
	
	Even if the CNN model already captures the most salient features by employing global max-pooling operations on multiple feature maps, the effectiveness of extraction depends not only on the number of feature maps but also on the initialisation of their filter weights. Usually, all the filter weights of a convolutional layer are initialised with small values, following a uniform distribution. The max-pooling operations can generate a smaller number of different features than the given number of feature maps as a result of using the same input and of applying filters with weights of the same distribution. Our approach to control better features extraction and to force the network to look for features in different regions of a text consists of applying the tiled CNN (TCNN), a variant of the CNN architecture which considers multiple filters per feature map. 
	
	\indent The TCNN is an improved CNN that was proposed for image processing to capture a wide range of picture invariances \cite{ngiam2010tiled}. Traditionally, the CNN architecture already learns the translational invariance due to its convolving operation but the TCNN goes forward and handles more complex cases such as viewpoint/rotation and size invariances. Since its development, the TCNN model has been mainly used for computer vision in tasks like image classification \cite{wang2015encoding}, object and emotion recognition \cite{gao2016local, qiu2018emotion}. 
	
	The reason behind the TCNN’s restricted applicability is its nature of multiple invariances learner. In this paper we adjust the model to meet the NLP requirements and set each filter of the feature map to cover only a set of $n$-grams (not all $n$-grams as in the case of pure CNN models). Following a precise filter structure per feature map, the extraction of the most relevant features is more effective and depends less on the initial values of the weights.
	
	In addition to the TCNN, a new network called hybrid tiled CNN (HTCNN) is introduced to outweigh the disadvantage of the TCNN model's inflexible filter structure. Using the idea of more diverse feature extraction with multiple filters, we create word clusters and allow a filter of a feature map to convolve only on the words that appear in the same context (assigned to the same cluster) and on their $n – 1$ neighbouring words (where $n$ is the size of the $n$-gram or the filter size). 
	
	In this paper, word clusters are computed using the expectation-maximization (EM) algorithm using Gaussian distribution. Since we do not include information about aspects, this approach allows us to identify only the document-level sentiment polarities.  However, HTCNN’s structure with multiple filters per feature map can be easily adjusted to ABSA sentiment classification task. An interesting idea could be to replace general word clusters defined a priori with sentence-level word clusters defined with respect to each aspect based on attention scores. We let this extension for future work.
	
	Our contributions are summarized as follows:
	\begin{enumerate}[noitemsep]
		\vspace{-1.3mm}
		\item We adjust the TCNN (which so far has been used only for image processing) to the NLP field.
		\item We design a novel hybrid tiled CNN model with multiple filters per feature map. The filter structure is flexible and each filter of the feature map can convolve only on the similar words of a given word cluster and on its $n-1$ neighbouring words, where $n$ is the window size. 
		\item Experimental results prove the effectiveness of the introduced model over the simple architecture of the CNN.
		%\vspace{-5mm}
	\end{enumerate}
	
	The remainder of the paper is organized as follows. The next section discusses the related literature. The third section depicts the TCNN in NLP and its hybrid variant. The fourth section presents the details of the experiments and the results and the last one concludes the paper. The source code used to implement the proposed models can be found at \url{https://github.com/mtrusca/HTCNN} 
	
	%\vspace{-3mm}
	\begin{figure*}[htb]
		\vspace{-0.2cm}
		\centering
		{\epsfig{file = 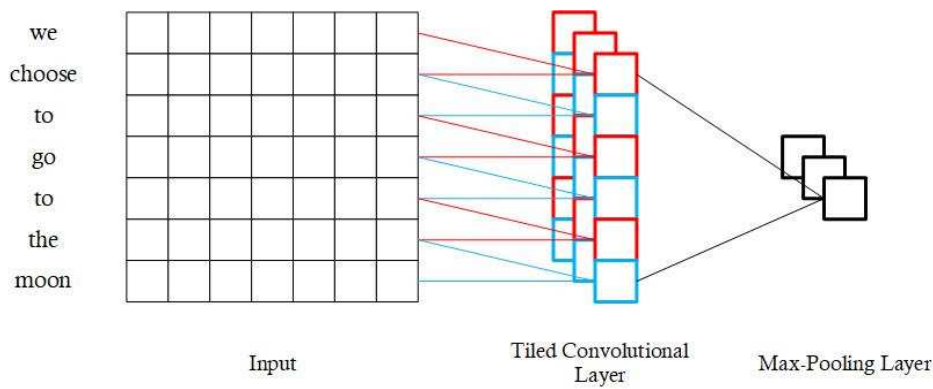, width = 0.8\linewidth}}
		\caption{The TCNN has three feature maps and each one has two filters convolving over bigrams. A global max-pooling operation is applied over the convolving results. The figure is adapted from \cite{kim2014convolutional}.}
		\label{fig:figure1}
		\vspace{-0.1cm}
	\end{figure*}	
	
	%\vfill
	
	\section{\uppercase{Related Work}}
	
	\noindent In text classification and sentiment analysis, CNN and RNN are the two main deep neural network architectures that produce similar results. While the CNN extracts the most relevant $n$-grams, the RNN finds context dependencies within an entire sentence. Still, for sentence classification, especially sentiment analysis it is tempting to use CNN models since polarities are usually determined only by some keywords \cite{yin2017comparative}. Regarding the sentence’s length, both neural networks have similar performances, especially for the case of short text. While for long sentences, the RNN could be considered more adequate than the CNN, some studies prove the contrary \cite{adel2016exploring, wen2016learning}, arguing that the RNN is a biased model where the later words of a sequence are more important than the others. 
	
	Currently, the CNN is a popular method for text classification \cite{kalchbrenner2014convolutional, hassan2018convolutional} and even models with little tuning of the hyperparameters provide competitive results \cite{kim2014convolutional, zhang2015sensitivity}. The variant of the CNN applied for sentiment analysis task could vary from simple approaches \cite{liao2017cnn} to more complex ones. For example, in \cite{yang2017overcoming}, the idea of linguistic homophily is exploited and sentiment polarities are found by combining a single-layer CNN model with information about users’ social proximity. In \cite{shin2016lexicon}, the results of a CNN sentiment model are improved by employing lexicon embeddings and attention mechanism. 
	
	Regarding the ABSA task, CNNs are not as popular as memory networks like LSTM and GRU even if they have shown potentiality to address this problem recently. The CNN model was introduced to tackle an ABSA problem in \cite{huang2019parameterized} by using parameterized gates and parameterized filters to include aspect information. Similar, in \cite{xue2018aspect}, it was proved that a simple CNN model with Gated Tanh-ReLU units is more accurate than the traditional approach with LSTM and attention mechanism. 
	
	Since not all datasets provide information about aspects, we try to increase the sensitivity of convolutional neural networks to the aspects and their related keywords by using a novel hybrid tiled CNN model. Our approach is similar to the one proposed in \cite{zhang2016mgnc} in terms of multiple inputs for simultaneous convolutional layers but instead of using different word embeddings we create word clusters and define sentence representation for each one.
	
	\section{\uppercase{Models}}
	
	\noindent The standard one-dimensional CNN architecture employed in \cite{collobert2011natural} and since then widely used in many papers, assumes that the feed-forward neural network is a composition of convolutional layers with the max-pooling operation and one or more fully connected layers with non-linear activation functions. An input text of length $m$ is represented as a matrix obtained by concatenating word embedding vectors \cite{kim2014convolutional}:
	
	\begin {equation}\label{equation1}
	X_{i:m} = \{w_1, w_2, ..., w_m\}
	\end {equation}
	
	This matrix is used as an input for a convolution operation that applies a filter $w$ on all possible windows of $n$ consecutive words at a given stride value. A new feature map is computed by applying a non-linear function $f$ with a bias term $b$ ($b$ $\in\mathbb{R}$) on the convolving results. Given the window of n words $X_{i:i+n-1}$, the generic object $C_i$ of the feature map is generated by:
	
	\begin {equation}\label{equation2}
	C_i = f(w{\ast}X_{i:i+n-1} + b)
	\end {equation}
	
	To capture the most important feature $C_{i}$, we apply the max-pooling operation.
	
	\begin {equation}\label{equation3}
	C_{max} = max\{C_1, C_2, ..., C_{m-n+1}\}
	\end {equation}
	
	Usually, a single feature is not enough for depicting the entire sentence and therefore it is a common practice to generate multiple feature maps for each convolutional layer. All features, with the highest value for each feature map, input the following convolutional or fully connected layers. In terms of sentiment analysis, we are interested to extract not only sentiment information but also information about text corpus’ entities and their aspects (attributes) \cite{liu2012sentiment}. Based on this need, the use of multiple feature maps per convolutional layer turns out to be a compulsory step. 

	\begin{figure*}[htb]
		\centering
		{\epsfig{file = 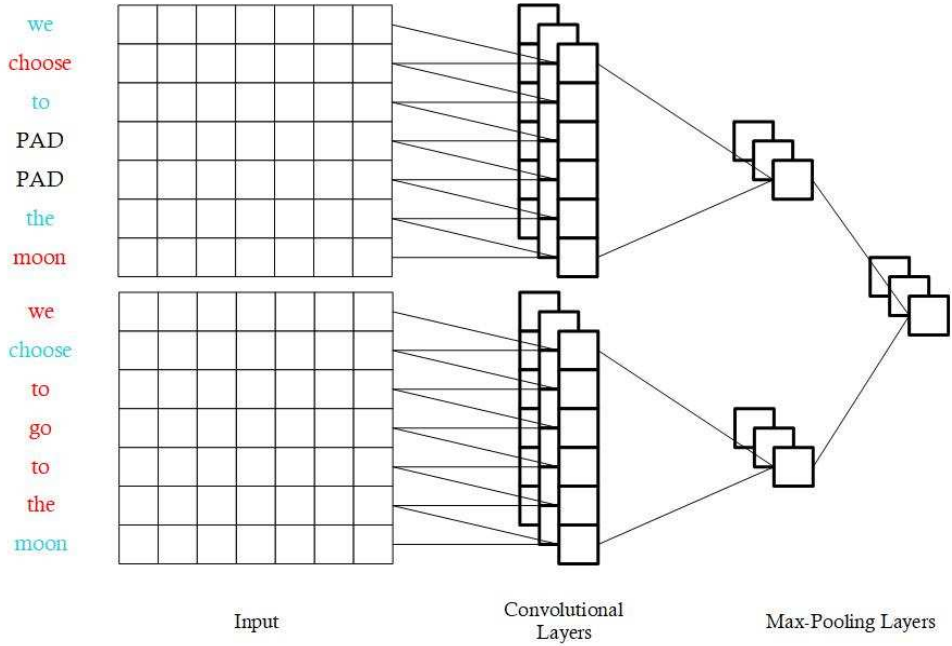, width = 0.8\linewidth}}
		\caption{HTCNN represented as a multiple simultaneous CNN with the number of clusters and the filter size equal to two and the number of feature maps equal to three.}
		\label{fig:figure2}
	\end{figure*}
	
	\subsection{TCNN}
	
	In this paper, we attempt to introduce the TCNN in the field of textual classification. The idea behind the TCNN is simple and assumes that each feature map is computed using at least two filters \cite{ngiam2010tiled}. The map of filters applied on the input is not randomly chosen but it has to follow a structure that imposes for each filter to have different nearby filters, while the next and previous $k$-th filters have to be the same. Term $k$ is known as the tile size and it is a hyperparameter that indicates the number of filters corresponding to each feature map. The generic object $C_i$ of a feature map is defined as:
	\begin {equation}\label{equation4}
	C_i = f(w_k{\ast}X_{i:i+n-1} + b)
	\end {equation}
	
	If $k$ is $1$ then the tiled convolutional layer turns into a simple convolutional layer and if $k$ is greater or equal to $m-n+1$ then we have a fully connected layer. While one of CNN's rules is parameter sharing, the tiled CNN follows it only partially so that weights are not entirely tied.    
	
	To better understand the way the tiled CNN works with textual data we consider an example sentence (“We choose to go to the moon”) in Figure \ref{fig:figure1}. We set the number of feature maps to three and the filter size and the tile size to two. The first filter convolves over the first two words ("we" and "choose"), then the second one covers the second and the third words ("choose" and "to"), then the first filter goes over the third and the fourth words ("to" and "go") and so on. This rule is applied for each feature map, which means that six weight matrices have to be initialised. Then a global max-pooling operation covers each of the three feature maps and the most representative $n$-grams are obtained (no matter the initial filter that identified them). The CNN's representation is similar to the one presented in Figure \ref{fig:figure1} but, unlike its tiled variant, there is only a filter per feature map (the tile size is equal to 1).
	
	The TCNN implementation can be seen as a neural network with multiple simultaneous convolutional layers.  Each layer corresponds to a filter and uses a stride value equal to the tile size. An important note to this process is related to the shape of the input that feeds each convolutional layer. Generally, filter $i$ ($i \leq k$) can slide from the $i$-th word to the word with the index equal to:
	%\begin {equation}\label{equation5}
	%\Big\lfloor \frac {m-n+1}{k} \Big\rfloor{\ast} k + (n-1) + \varphi_i
	%end {equation}
	%\begin {equation}
	%\label{equation6}
	%\varphi_i = \begin {cases} 
	%{i},  & \text{if } \frac {i}{k} {\leq} \{\frac{m-n+1}{k}\}\\
	%{k - i},       &\text{if } \frac {i}{k} > \{\frac{m-n+1}{k}\}\\
	%\end {cases}
	%\end {equation}

	\begin {equation}\label{equation5}
	\Big\lfloor \frac {m}{k + n -1} \Big\rfloor{\ast} (k + n - 1) + \varphi_i
	\end {equation}
	\begin {equation}
	\label{equation6}
	\varphi_i = \begin {cases} 
	{(n + i - 2},  & \text{if } \{\frac{m}{k + n -1}\}   {\geq} \frac{i}{k}\\
	{i - k}, &\text{if } \{\frac{m}{k + n -1}\} < \frac{i}{k}\\
	\end {cases}
	\end {equation}
	
	\noindent {where $\lfloor x \rfloor$ and $\{ x \}$ represent the integer and decimal part of a number $x$, respectively. According to this rule, we have to adjust the input of each convolutional layer. This rule together with the constraint of the stride value enforces the filters to convolve on different $n$-grams and to cover together the entire input. Before moving further, we have to add two sets of pooling layers. The layers of the first set correspond to the $k$ convolutional layers. The second set has just one layer covering the previous concatenated results. By this means, it is assured that the multiple simultaneous CNN behaves just like a TCNN model with a single pooling operation.}
	
	The reasoning behind using the TCNN model for text classification is the need to capture more diverse features (like aspects and sentiment words) that could take our results closer to the sentences’ labels. Even if this task is theoretically accomplished by the simple CNN, the risk of getting a smaller number of different features than the number of feature maps is high. On the other side, TCNN with its integrated multiple filters per feature map forces the model to look for features in diverse regions of a sequence enhancing the quality of extraction. By choosing different values of $k$, we get a palette of models that are a trade-off between estimating a small number of parameters and having more different features.

	\begin{table*}[h]
		\caption{The configuration of models for IMDB and SemEval 2017 datasets}\label{tab:table4} \centering
		\begin{spreadtab}{{tabular}{|c|c|c|c|}}
			\hline
			@\multicolumn{2}{|c|}{{\text{}}} &
			@\multicolumn{1}{|c|}{\text{IMDB}} &
			@\multicolumn{1}{|c|}{\text{SemEval 2017}}\\
			\hline
			@\multicolumn{1}{|c|}{\multirow{3}{*}{\text{Activation Function}}} &
			@\multicolumn{1}{|c|}{\text{Convolutional Layer}} &
			@\multicolumn{1}{|c|}{\text{ReLu}} &
			@\multicolumn{1}{|c|}{\text{ReLu}}\\
			\cline{2-4}
			@\multicolumn{1}{| c |}{\text{}} 
			& @\text{Fully Connected Layer} 
			& @\text{Tanh} 
			& @\text{ReLu}\\
			\cline{2-4}
			@\multicolumn{1}{| c |}{\text{}} 
			& @\text{Output Layer} 
			& @\text{Sigmoid} 
			& @\text{Softmax}\\
			\hline
			@\multicolumn{2}{|c|}{{\text{Filter Size (n)}}} &
			@\multicolumn{1}{|c|}{\text{2, 3, 4}} &
			@\multicolumn{1}{|c|}{\text{2, 3, 4}}\\
			\hline
			@\multicolumn{2}{|c|}{{\text{Tile Size / No. of Clusters (k)}}} &
			@\multicolumn{1}{|c|}{\text{2, 3}} &
			@\multicolumn{1}{|c|}{\text{3}}\\
			\hline
		\end{spreadtab}
	\end{table*}
	
	\subsection{HTCNN}
	
	The disadvantage of TCNN is the inflexible structure of filters sliding on the input that requires to have $k$ repetitive filters with no possibility to change the ordering. Knowing that words with semantic similarity are likely to appear in a similar context \cite{harris1954distributional}, one reasonable filter map could be the one that applies the same filter on all these words and other filters on the remaining words. For example in a sentence like “The lens of my digital camera have a powerful/weak optical zoom” it could be useful to apply the same filter on the “powerful” and “weak” words and different ones on the other words.
	
	Knowing that word embeddings are built to capture not only words’ context but also their syntactic and semantic similarity, words like “powerful” and “weak” have similar word embeddings and could be assigned to the same cluster. Using an appropriate clustering model, we can group the word vectors in $k$ clusters. In the experiments, the clustering solution is provided by the EM algorithm using Gaussian distribution. The reason behind this choice is that Gaussian mixture models support mixed membership and are robust against errors and flexible regarding cluster covariance \cite{meila2013experimental}. The EM algorithm has two steps that are repeatedly applied till the convergence is noticed. First, we compute the probability of a word to belong to a cluster, knowing that the words of clusters follow normal distributions (the expectation step). Second, we update the cluster means accordingly with the previously computed probabilities (the maximization step).
	
	For each cluster, we code the outside cluster words of a sentence with “PAD” and leave unchanged its associated words. Sequences generated based on indices of dictionary words, get 0 for “PAD” word and a value greater than 0 otherwise. In this way, we define $k$ inputs (one for each cluster) and use them to feed $k$ multiple simultaneous convolutional layers. The stride value constraint is used no more because simultaneous layers already convolve on different words. 
	
	The problem of the last depicted neural network is the loss of the $n$-grams concept caused by the replacement of the some words’ indexes with 0 (words assigned to other clusters than the one associated with the given convolutional layer). This idea is supported by the assumption that words are dependent on their neighbouring words. To overcome this problem, for each convolutional layer we have to modify the input by adding in the left and in the right side (if it is possible) of each word (whose index is different than 0) its $n-1$ neighbouring words. The added words could be in the same cluster or not. We call this model, the hybrid tiled CNN (HTCNN).
	
	Considering the above sentence and setting the filter size and the number of clusters to two (the same as setting the tile size value for the TCNN), we assign the words “choose” and “moon” to the first cluster and the other words to the second one. Firstly we assign to the each cluster the same sentence and replace the words of the other cluster with “PAD”. The modified input sentences are: “PAD choose PAD PAD PAD PAD moon” and “we PAD to go to the PAD”. If we add the neighbors (a single word to the left and the right for bigram cases) of each word, the input sentences will change into “we choose to PAD PAD the moon” and “we choose to go to the moon”. The sentence of the first cluster gets the words “we” and “to” as the neighbors of the “choose” word and the word “the” as the neighbor of the “moon” word. The words of the second cluster are: “we”, “to”, “go”, “to”, “the” and the sentence gets the word “choose” as the neighbor of the words “we” and “to” and the word “moon” as the neighbor of the word “the”. This process is described in Figure \ref{fig:figure2}, for representative reasons we use sentences instead of sequences. The colour of the cluster words assigned to each convolutional layer is red. The colour of the neighbouring words is blue. The remaining words are coded with the word "PAD". If the blue words are replaced with "PAD" words then we get a new model that does not consider the $n$-gram concept and it is called simple HTCNN (SHTCNN).
	
	\section{\uppercase{Experimental Results}}
	
	\noindent We test the TCNN and the HTCNN models on the IMDB movie reviews dataset \cite{maas2011learning}. The collection has 50000 documents, evenly separated between negative and positive reviews. 
	
	%Our models based on convolution operation are compared to a one-layer one-dimensional CNN. 	We consider the following options for activation functions: ReLU and hyperbolic tangent functions for hidden layers and softmax and sigmoid functions for the output layer. In our case, the best performance of the simple CNN is achieved using the following activation functions: ReLU for the convolutional layer, hyperbolic tangent for fully connected layer and sigmoid for the output layer. As CNN is the reference model, we apply the same activation functions to all presented models. Regarding word vectors initialisation, we use word embeddings trained on the current dataset with word2vec models \cite{mikolov2013efficient}. Each word is represented by a 200 dimension vector computed by concatenating a continuous bag-of-words (CBOW) 100 dimension vector with a skip-gram (SG) vector of the same size. Sentences are represented using zero-padding strategy which means we are applying the wide convolutions \cite{kalchbrenner2014convolutional}.
	
	Our models based on convolution operation are compared to a one-layer one-dimensional CNN. The configuration of the CNN baseline model and of the proposed models is presented in Table \ref{tab:table4}. Regarding word vectors initialisation, we use word embeddings trained on the current dataset with word2vec models \cite{mikolov2013efficient}. Each word is represented by a 200 dimension vector computed by concatenating a continuous bag-of-words (CBOW) 100 dimension vector with a skip-gram (SG) vector of the same size. Sentences are represented using zero-padding strategy which means we are applying the wide convolutions \cite{kalchbrenner2014convolutional}.
	
	Knowing that the IMBD dataset is balanced, the evaluation of models is done in terms of test accuracy. We use the binary cross entropy loss function and unfreeze the word embeddings during the training phase. Using 4-fold cross-validation, 10,000 reviews are allocated to the testing set and the rest of them are divided between training and validation sets. The cross-validation process has four iterations and at each time step other 10,000 reviews are allocated to the validation set. %We test the models considering the filter size equal to two, three and four (bigrams, trigrams and fourgram) and the tile size/number of clusters equal to two and three. The words are represented only by the 100 dimension skip-gram vectors (for computational cost reduction). 
	
	In addition to the TCNN and its hybrid variant, we present the results SHTCNN model to see how the lack of the neighbouring words affects the performance of the HTCNN. The results of our comparison between the baseline network and our models are listed in Table \ref{tab:table1}.
	
	The baseline CNN has a volatile performance, unlike all other models. While for a filter size equal to two the model has one of the best test accuracies, for bigger sizes (trigram and fourgram cases) the performance decreases quickly and reaches the lowest threshold. All proposed networks have a natural behaviour and their performance grows proportionally with the filter size. Besides the $n$-gram dimension, the tile size or the number of word clusters has also impact on the results, generally leading to a gradual improvement.
	
	Results of the TCNN and of the SHTCNN are complementary and while the first one performs better when we set the tile size to three, the second one has better results for two word clusters. However, we note that the HTCNN achieves the best performance without having the drawbacks of the other two models: the inflexible filter structure of the the TCNN and the SHTCNN’s lack of $n$-grams. 
	
	Natural, more filters per feature map makes the model more sensitive to different features but in the same way, increases the number of estimated parameters or the model complexity. Further on, the HTCNN is tested for larger tile sizes, precisely for four and five word clusters. The results are listed in Table \ref{tab:table2}. Only the pair (k = 4, n = 2) has a slightly better result than the previous model variants with the same filter size (the pairs (k = 3, n = 2) and (k = 2, n = 2)). All other results are worse than the ones of HTCNN with the tile size equal to three and the difference is more significant for the case of five word clusters. We conclude that the optimal tile size value corresponding to the IMBD dataset that balances the model capacity to extract different features with its complexity is equal to three.
	
	To confirm the performance of HTCNN over single-layer CNN, we undertake a second test on the SemEval-2017 English dataset for task 4: Sentiment analysis in Twitter, subtask A: Message Polarity Classification \cite{rosenthal2017semeval}. The dataset has 62,617 tweets with the following statistics: 22,277 positive tweets, 28,528 neutral tweets and 11,812 negative tweets. The approach with the highest rank on this subtask was proposed by Cliche \cite{cliche2017bb_twtr} and ensembles a LSTM and a neural network with multiple simultaneous convolutional layers. Since the Cliche’s CNN structure is similar to the structure of HTCNN, we use some of his default settings: the length of the word embeddings, the activation functions, the number of filters per convolutional layer, the dropout layers (with the same dropout probabilities) and the number of epochs. The number of simultaneous convolutional layers utilizes in the Cliche’s structure is equal to three which means that we have to run our HTCNN model for the same number of filters. Table \ref{tab:table4} shows an overview of the models' architecture for the SemEval dataset.  
	
	Word embeddings are trained using CBOW and SG models in a similar way as the one depicted above for the IMDB dataset and remain frozen over the entire training phase. 10\% of the corpus tweets are allocated to the testing set while the remaining tweets are separated between validation and training sets using 9-fold cross-validation. Because the dataset is imbalanced, the evaluation is done using macro-average testing recall. To counteract the imbalance, the cross-entropy loss function uses class weights adjusted inversely proportional to the class frequencies in the training phase. Both the HTCNN and the CNN share the above hyperparameters. Similar to the IMDB experiments, we set the window size to two, three or four. The results of our comparison on the SemEval dataset are shown in Table \ref{tab:table3} and confirm the performance of HTCNN over simple CNN.

	\begin{table*}[h]
		\caption{Comparison between CNN and TCNN, SHTCNN and HTCNN on the IMDB dataset. (\textbf{Bold} denotes the best performance per model and \textbf{italics} the best performance overall)}\label{tab:table1} \centering
		\begin{spreadtab}{{tabular}{|c|c|c|c|c|c|c|c|}}
			\hline
			@\multicolumn{1}{| c |}{\multirow{4}{*}{\text{}}} &
			@\multicolumn{1}{|c|}{\text{CNN}} &
			@\multicolumn{2}{|c|}{\text{TCNN}} &
			@\multicolumn{2}{|c|}{\text{SHTCNN}} &
			@\multicolumn{2}{|c|}{\text{HTCNN}}\\
			%@\multicolumn{1}{| c |}{\text{}} &
			%@\multicolumn{1}{|c|}{\text{CNN}} &
			%@\multicolumn{2}{|c|}{\text{}} &
			%@\multicolumn{2}{|c|}{\text{tiled CNN}} &
			%@\multicolumn{2}{|c|}{\text{CNN}}\\
			\cline{2-8}
			@\multicolumn{1}{| c |}{\text{}} 
			& @\text{accuracy} 
			& @\text{accuracy} 
			& @\text{p-value*}
			& @\text{accuracy} 
			& @\text{p-value*}
			& @\text{accuracy} 
			& @\text{p-value*}\\
			\hline
			@\text{k=2**; n=2}
			& @\text{\text{0.8783}} & @\text{\textbf{0.8789}} & @\text{0.8560} & @\text{0.8641} & @\text{0.3739} & @\text{0.8688} & @\text{0.0410} \\
			\hline
			@\text{k=2**; n=3}
			& @\text{0.8445} & @\text{0.8586} & @\text{0.2677} & @\text{0.8667} & @\text{0.1203} & @\textbf{0.8718} &	 @\text{0.0720} \\
			\hline
			@\text{k=2**; n=4}
			& @\text{0.8528} & @\text{0.8577} & @\text{0.4685} & @\text{\textbf{0.8786}} & @\text{0.0158} & @\text{0.8723} & @\text{0.0387} \\
			\hline
			@\text{k=3**; n=2}
			& @\text{\text{0.8783}} & @\text{0.8693} & @\text{0.0610} & @\text{0.8658} & @\text{0.3820} & @\textbf{0.8786} & @\text{0.9360} \\
			\hline
			@\text{k=3**; n=3}
			& @\text{0.8445} & @\text{0.8775} & @\text{0.0191} & @\text{0.8727} & @\text{0.0439} & @\textbf{0.8796} & @\text{0.0166} \\
			\hline
			@\text{k=3**; n=4}
			& @\text{0.8528} & @\text{0.8762} & @\text{0.0113} & @\text{0.8736} & @\text{0.0360} & @\text{\textbf{\textit{0.8825}}} & @\text{0.0025} \\
			\hline
		\end{spreadtab}
		
		\justify
		\noindent {\footnotesize * P-value is the probability of the null hypothesis associated with the paired Student’s t-test that measures the similarity between the baseline model and each of the three networks.}
		
		\noindent {\footnotesize ** The number of clusters is equal to the tile size (k).}
	\end{table*}
	
	\begin{table}[h]
		\caption{Comparison between CNN and HTCNN on the IMDB dataset for larger tile sizes. (\textbf{Bold} denotes the best performance per model and \textbf{italics} the best performance overall)}\label{tab:table2} \centering
		\begin{spreadtab}{{tabular}{|c|c|c|c|}}
			\hline
			@\multicolumn{1}{| c |}{\multirow{4}{*}{\text{}}} &
			@\multicolumn{1}{|c|}{\text{CNN}} &
			@\multicolumn{2}{|c|}{\text{HTCNN}}\\
			%@\multicolumn{1}{| c |}{\text{}} &
			%@\multicolumn{1}{|c|}{\text{CNN}} &
			%@\multicolumn{2}{|c|}{\text{CNN}}\\
			\cline{2-4}
			@\multicolumn{1}{| c |}{\text{}} 
			& @\text{accuracy} 
			& @\text{accuracy} 
			& @\text{p-value}\\
			\hline
			@\text{k=4; n=2}
			& @\text{\text{0.8783}} & @\textbf{0.8788} & @\text{0.8719} \\
			\hline
			@\text{k=4; n=3}
			& @\text{0.8445} & @\textbf{0.8674} & @\text{0.0699} \\
			\hline
			@\text{k=4; n=4}
			& @\text{0.8528} & @\text{\textbf{\textit{0.8815}}} & @\text{0.0010} \\
			\hline
			@\text{k=5; n=2}
			& @\text{\textbf{0.8783}} & @\text{0.8767} & @\text{0.6096} \\
			\hline
			@\text{k=5; n=3}
			& @\text{0.8445} & @\textbf{0.8758} & @\text{0.0236} \\
			\hline
			@\text{k=5; n=4}
			& @\text{0.8528} & @\textbf{0.8812} & @\text{0.0025} \\
			\hline
		\end{spreadtab}
	\end{table}

	\begin{table}[h]
		\caption{Comparison between CNN and HTCNN on the SemEval-2017 dataset, task 4: Sentiment analysis in Twitter, subtask A: Message Polarity Classification. (\textbf{Bold} denotes the best performance per model and \textbf{italics} the best performance overall)}\label{tab:table3} \centering
		\begin{spreadtab}{{tabular}{|c|c|c|c|}}
			\hline
			@\multicolumn{1}{| c |}{\multirow{4}{*}{\text{}}} &
			@\multicolumn{1}{|c|}{\text{CNN}} &
			@\multicolumn{2}{|c|}{\text{HTCNN}}\\
			%@\multicolumn{1}{| c |}{\text{}} &
			%@\multicolumn{1}{|c|}{\text{CNN}} &
			%@\multicolumn{2}{|c|}{\text{CNN}}\\
			\cline{2-4}
			@\multicolumn{1}{| c |}{\text{}} 
			& @\text{recall} 
			& @\text{recall} 
			& @\text{p-value}\\
			\hline
			@\text{k=3; n=2}
			& @\text{0.7285} & @\textbf{0.7715} & @\text{0.1783} \\
			\hline
			@\text{k=3; n=3}
			& @\text{0.7881} & @\textbf{\textbf{\textit{0.8516}}} & @\text{0.0400} \\
			\hline
			@\text{k=3; n=4}
			& @\text{\text{0.8042}} & @\textbf{0.8472} & @\text{0.0425} \\
			\hline
		\end{spreadtab}
	\end{table}	
	
	\section{\uppercase{Conclusions}}
	
	\noindent In this paper, we present the TCNN model in the field of sentiment classification to improve the process of feature extraction. Due to the constraint of partial untying of filter weights that forces the network to apply the same filter at each k step, we introduced the HTCNN model. The new architecture combines the benefits of word embeddings clustering with the idea of tiling and if the tile size is chosen properly, the model could achieve competitive performance with the TCNN and better results than the reference CNN model. Setting the tile size or the number of word clusters could be difficult. Because the HTCNN works like a simultaneous CNN structure, too many parallel convolutional layers could lead to both higher feature sensitivity and higher complexity. CNN does not implicitly fit the fine-grained aspect based sentiment analysis, but using the right network configuration we can adjust it to be more sensitive to the corpus’ features and to increase the overall performance. Even if the results of our experiments are modest compared with the state-of-art sentiment analysis models for IMDB and SemEval-2017 datasets, we prove that the HTCNN is a good substitute for CNN in the NLP field and the replacement of the CNN with the HTCNN in more laborious architectures could lead to higher rates of performance than CNN.
	
	As our future work, it would be interesting to see how word networks based on word embeddings improve the clustering solution incorporated in the HTCNN, perhaps in an approach similar to the one proposed in \cite{feria2018constructing}. Moreover, a promising direction would be to have meaningful clusters that denote specific functions (e.g. aspect) or other roles (e.g. syntactic). This would give a boost to the transparency of deep models in natural language processing, since the CNN filters (and consequently, the weights) would be directly interpretable. 
	
	\bibliographystyle{apalike}
	{\small
		\bibliography{paper_bibliography}}

\begin{thebibliography}{}

\bibitem[Adel and Sch{\"u}tze, 2016]{adel2016exploring}
Adel, H. and Sch{\"u}tze, H. (2016).
\newblock Exploring different dimensions of attention for uncertainty
  detection.
\newblock {\em arXiv preprint arXiv:1612.06549}.

\bibitem[Cliche, 2017]{cliche2017bb_twtr}
Cliche, M. (2017).
\newblock Bb\_twtr at semeval-2017 task 4: Twitter sentiment analysis with cnns
  and lstms.
\newblock {\em arXiv preprint arXiv:1704.06125}.

\bibitem[Collobert et~al., 2011]{collobert2011natural}
Collobert, R., Weston, J., Bottou, L., Karlen, M., Kavukcuoglu, K., and Kuksa,
  P. (2011).
\newblock Natural language processing (almost) from scratch.
\newblock {\em Journal of machine learning research}, 12(Aug):2493--2537.

\bibitem[Feng et~al., 2015]{feng2015applying}
Feng, M., Xiang, B., Glass, M.~R., Wang, L., and Zhou, B. (2015).
\newblock Applying deep learning to answer selection: A study and an open task.
\newblock In {\em 2015 IEEE Workshop on Automatic Speech Recognition and
  Understanding (ASRU)}, pages 813--820. IEEE.

\bibitem[Feria et~al., 2018]{feria2018constructing}
Feria, M., Balbin, J.~P., and Bautista, F.~M. (2018).
\newblock Constructing a word similarity graph from vector based word
  representation for named entity recognition.
\newblock {\em arXiv preprint arXiv:1807.03012}.

\bibitem[Gao and Lee, 2016]{gao2016local}
Gao, Y. and Lee, H. (2016).
\newblock Local tiled deep networks for recognition of vehicle make and model.
\newblock {\em Sensors}, 16(2):226.

\bibitem[Harris, 1954]{harris1954distributional}
Harris, Z.~S. (1954).
\newblock Distributional structure.
\newblock {\em Word}, 10(2-3):146--162.

\bibitem[Hassan and Mahmood, 2018]{hassan2018convolutional}
Hassan, A. and Mahmood, A. (2018).
\newblock Convolutional recurrent deep learning model for sentence
  classification.
\newblock {\em Ieee Access}, 6:13949--13957.

\bibitem[Huang and Carley, 2019]{huang2019parameterized}
Huang, B. and Carley, K.~M. (2019).
\newblock Parameterized convolutional neural networks for aspect level
  sentiment classification.
\newblock {\em arXiv preprint arXiv:1909.06276}.

\bibitem[Kalchbrenner et~al., 2014]{kalchbrenner2014convolutional}
Kalchbrenner, N., Grefenstette, E., and Blunsom, P. (2014).
\newblock A convolutional neural network for modelling sentences.
\newblock {\em arXiv preprint arXiv:1404.2188}.

\bibitem[Kim, 2014]{kim2014convolutional}
Kim, Y. (2014).
\newblock Convolutional neural networks for sentence classification.
\newblock {\em arXiv preprint arXiv:1408.5882}.

\bibitem[Lai et~al., 2015]{lai2015recurrent}
Lai, S., Xu, L., Liu, K., and Zhao, J. (2015).
\newblock Recurrent convolutional neural networks for text classification.
\newblock In {\em Twenty-ninth AAAI conference on artificial intelligence}.

\bibitem[LeCun et~al., 1999]{lecun1999object}
LeCun, Y., Haffner, P., Bottou, L., and Bengio, Y. (1999).
\newblock Object recognition with gradient-based learning.
\newblock In {\em Shape, contour and grouping in computer vision}, pages
  319--345. Springer.

\bibitem[Liao et~al., 2017]{liao2017cnn}
Liao, S., Wang, J., Yu, R., Sato, K., and Cheng, Z. (2017).
\newblock Cnn for situations understanding based on sentiment analysis of
  twitter data.
\newblock {\em Procedia computer science}, 111:376--381.

\bibitem[Liu, 2012]{liu2012sentiment}
Liu, B. (2012).
\newblock Sentiment analysis and opinion mining.
\newblock {\em Synthesis lectures on human language technologies}, 5(1):1--167.

\bibitem[Liu and Zhang, 2012]{liu2012survey}
Liu, B. and Zhang, L. (2012).
\newblock A survey of opinion mining and sentiment analysis.
\newblock In {\em Mining text data}, pages 415--463. Springer.

\bibitem[Ma et~al., 2018]{ma2018targeted}
Ma, Y., Peng, H., and Cambria, E. (2018).
\newblock Targeted aspect-based sentiment analysis via embedding commonsense
  knowledge into an attentive lstm.
\newblock In {\em Thirty-Second AAAI Conference on Artificial Intelligence}.

\bibitem[Maas et~al., 2011]{maas2011learning}
Maas, A.~L., Daly, R.~E., Pham, P.~T., Huang, D., Ng, A.~Y., and Potts, C.
  (2011).
\newblock Learning word vectors for sentiment analysis.
\newblock In {\em Proceedings of the 49th annual meeting of the association for
  computational linguistics: Human language technologies-volume 1}, pages
  142--150. Association for Computational Linguistics.

\bibitem[Meila and Heckerman, 2013]{meila2013experimental}
Meila, M. and Heckerman, D. (2013).
\newblock An experimental comparison of several clustering and initialization
  methods.
\newblock {\em arXiv preprint arXiv:1301.7401}.

\bibitem[Mikolov et~al., 2013]{mikolov2013efficient}
Mikolov, T., Chen, K., Corrado, G., and Dean, J. (2013).
\newblock Efficient estimation of word representations in vector space.
\newblock {\em arXiv preprint arXiv:1301.3781}.

\bibitem[Mou et~al., 2015]{mou2015natural}
Mou, L., Men, R., Li, G., Xu, Y., Zhang, L., Yan, R., and Jin, Z. (2015).
\newblock Natural language inference by tree-based convolution and heuristic
  matching.
\newblock {\em arXiv preprint arXiv:1512.08422}.

\bibitem[Ngiam et~al., 2010]{ngiam2010tiled}
Ngiam, J., Chen, Z., Chia, D., Koh, P.~W., Le, Q.~V., and Ng, A.~Y. (2010).
\newblock Tiled convolutional neural networks.
\newblock In {\em Advances in neural information processing systems}, pages
  1279--1287.

\bibitem[Qiu et~al., 2018]{qiu2018emotion}
Qiu, J.-L., Qiu, X.-Y., and Hu, K. (2018).
\newblock Emotion recognition based on gramian encoding visualization.
\newblock In {\em International Conference on Brain Informatics}, pages 3--12.
  Springer.

\bibitem[Rosenthal et~al., 2017]{rosenthal2017semeval}
Rosenthal, S., Farra, N., and Nakov, P. (2017).
\newblock Semeval-2017 task 4: Sentiment analysis in twitter.
\newblock In {\em Proceedings of the 11th international workshop on semantic
  evaluation (SemEval-2017)}, pages 502--518.

\bibitem[Shin et~al., 2016]{shin2016lexicon}
Shin, B., Lee, T., and Choi, J.~D. (2016).
\newblock Lexicon integrated cnn models with attention for sentiment analysis.
\newblock {\em arXiv preprint arXiv:1610.06272}.

\bibitem[Thet et~al., 2010]{thet2010aspect}
Thet, T.~T., Na, J.-C., and Khoo, C.~S. (2010).
\newblock Aspect-based sentiment analysis of movie reviews on discussion
  boards.
\newblock {\em Journal of information science}, 36(6):823--848.

\bibitem[Wang and Liu, 2015]{wang2015deep}
Wang, B. and Liu, M. (2015).
\newblock Deep learning for aspect-based sentiment analysis.
\newblock {\em Stanford University report}.

\bibitem[Wang and Oates, 2015]{wang2015encoding}
Wang, Z. and Oates, T. (2015).
\newblock Encoding time series as images for visual inspection and
  classification using tiled convolutional neural networks.
\newblock In {\em Workshops at the Twenty-Ninth AAAI Conference on Artificial
  Intelligence}.

\bibitem[Wen et~al., 2016]{wen2016learning}
Wen, Y., Zhang, W., Luo, R., and Wang, J. (2016).
\newblock Learning text representation using recurrent convolutional neural
  network with highway layers.
\newblock {\em arXiv preprint arXiv:1606.06905}.

\bibitem[Xue and Li, 2018]{xue2018aspect}
Xue, W. and Li, T. (2018).
\newblock Aspect based sentiment analysis with gated convolutional networks.
\newblock {\em arXiv preprint arXiv:1805.07043}.

\bibitem[Yang and Eisenstein, 2017]{yang2017overcoming}
Yang, Y. and Eisenstein, J. (2017).
\newblock Overcoming language variation in sentiment analysis with social
  attention.
\newblock {\em Transactions of the Association for Computational Linguistics},
  5:295--307.

\bibitem[Yin et~al., 2017]{yin2017comparative}
Yin, W., Kann, K., Yu, M., and Sch{\"u}tze, H. (2017).
\newblock Comparative study of cnn and rnn for natural language processing.
\newblock {\em arXiv preprint arXiv:1702.01923}.

\bibitem[Zhang et~al., 2012]{zhang2012weakness}
Zhang, W., Xu, H., and Wan, W. (2012).
\newblock Weakness finder: Find product weakness from chinese reviews by using
  aspects based sentiment analysis.
\newblock {\em Expert Systems with Applications}, 39(11):10283--10291.

\bibitem[Zhang et~al., 2016]{zhang2016mgnc}
Zhang, Y., Roller, S., and Wallace, B. (2016).
\newblock Mgnc-cnn: A simple approach to exploiting multiple word embeddings
  for sentence classification.
\newblock {\em arXiv preprint arXiv:1603.00968}.

\bibitem[Zhang and Wallace, 2015]{zhang2015sensitivity}
Zhang, Y. and Wallace, B. (2015).
\newblock A sensitivity analysis of (and practitioners' guide to) convolutional
  neural networks for sentence classification.
\newblock {\em arXiv preprint arXiv:1510.03820}.

\end{thebibliography}
	
\end{document}